\documentclass[11pt]{article}
\usepackage[british]{babel}

\usepackage[final]{lrec2026}
\usepackage[T1]{fontenc}
\usepackage[utf8]{inputenc}
\usepackage{inconsolata}

\usepackage{graphicx} 
\usepackage{multirow} 
\usepackage{float}
\usepackage{url} 
\usepackage{tabularx} 
\usepackage{subfigure} 

\usepackage{microtype}
\usepackage{times}
\usepackage{latexsym}

\title{Finding Meaning in Embeddings: Concept Separation Curves}

\name{Paul Keuren$^{\ast}{}^{\dagger}$, Marc Ponsen $^{\dagger}$, Robert A. Bagheri$^{\ast}$}
\address{$^{\ast}$Utrecht University \\
          {$^{\dagger}$Statistics Netherlands}\\
          p.j.g.keuren@uu.nl}

\date{}
\abstract{
Sentence embedding techniques aim to encode key concepts of a sentence’s meaning in a vector space.
However, the majority of evaluation approaches for sentence embedding quality rely on the use of additional classifiers or downstream tasks.
These additional components make it unclear whether good results stem from the embedding itself or from the classifier's behaviour. 
In this paper, we propose a novel method for evaluating the effectiveness of sentence embedding methods in capturing sentence-level concepts. 
Our approach is classifier-independent, allowing for an objective assessment of the model's performance. 
The approach adopted in this study involves the systematic introduction of syntactic noise and semantic negations into sentences, with the subsequent quantification of their relative effects on the resulting embeddings. 
The visualisation of these effects is facilitated by Concept Separation Curves, which show the model's capacity to differentiate between conceptual and surface-level variations. 
By leveraging data from multiple domains, employing both Dutch and English languages, and examining sentence lengths, this study offers a compelling demonstration that Concept Separation Curves provide an interpretable, reproducible, and cross-model approach for evaluating the conceptual stability of sentence embeddings.
The code is open source and located on \href{https://github.com/pkun-cbs/ConceptSeparationCurves}{github}, and a live interactive demo is available at \href{https://conceptseparationcurves.streamlit.app/}{streamlit}.
 \newline \Keywords{sentence embedding, embedding evaluation, conceptual representation, concept separation curves, large language models}
}
 
\begin{document}

\maketitleabstract


\section{Introduction}
How can we be certain that a Large Language Model (LLM) is capable of distinguishing between concepts?
Oftentimes, this is tested by prompting an LLM and inspecting the agreement of the result~\cite{kiyak2024chatgpt,yetisensoy2025validity}.
The problems with this approach are twofold. 
Firstly, it is reliant on costly human annotations~\cite{wang-etal-2021-want-reduce}. 
Secondly, it makes the implicit assumption that correct answers necessarily reflect genuine understanding, even though models may succeed without representing the underlying concepts\cite{adiFinegrainedAnalysisSentence2017,bender2020climbing}.
The annotations become even more difficult when such an answer cannot be evaluated by hand, as is the case with the output from sentence encoders~\cite{attention_is_all_you_need}.

\begin{figure}
    \centering
    \includegraphics[width=\linewidth]{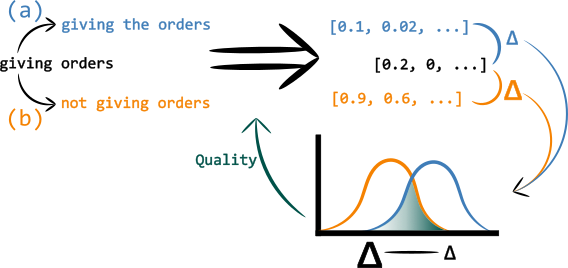}
    \caption{
    Concept Separation Curves. This example has been translated from the Dutch sentence "bevelen geven" (giving orders), which originates from the CompetentNL dataset. Initially, a set of perturbations is computed for a given sentence: a) surface-level perturbation, and b) semantic change. Following the process of embedding each sentence, the difference per vector is measured. The application of this process to the entire corpus provides insight into the quality of the embedding, as demonstrated by the overlap between the curves.
    }
    \label{fig:summary}
    \vspace{-10pt}
\end{figure}

In this research, we propose a method which resolves the annotation issue and sheds light on the understanding.
Our method, summarised in Figure~\ref{fig:summary}, is applied to different sentence embedding methods.
By taking a corpus and performing perturbations, we can derive a measure of concept separability by the embedding method.
The perturbations are key, as they are required to embody different concepts.
A concept is defined as a thought or idea~\cite{vocabulary_concept_2025}, and as such, it can be difficult to generate sentences with different concepts automatically.
Due to this, we define a set of rules for the perturbation and implement an example in the form of negation.

Unlike existing evaluation approaches(\cite{adiFinegrainedAnalysisSentence2017}), our method does not rely on annotated datasets or classifiers; instead, it isolates semantic changes (such as negations) from surface-level perturbations (such as added noise). 
This makes it possible to directly assess how sensitively an embedding represents meaning, independent of external modelling choices. 
Although we use our approach on data in two languages (Dutch and English), it is designed to generalise across others as well.

To make these claims concrete, we make the following contributions:
\begin{itemize}
    \item We introduce \textbf{Concept Separation Curves (CSCs)}, a classifier-independent method for evaluating sentence embeddings by contrasting semantic and surface-level perturbation.
    \item We propose \textbf{an automated, annotation-free perturbation framework} based on controlled alteration.
    \item We define a geometric overlap measure to quantify separation between semantic and non-semantic effects in embedding space.
    \item We analyse embedding behaviour across models, languages, and sentence lengths, identifying failure modes related to token position sensitivity and sentence length.
\end{itemize}

In the rest of the paper, we will delve into related work. 
Then, we describe our methods, which focus on language analysis instead of expert knowledge. 
Finally, we discuss the results and implications of our methods.

\section{Related work}
To give a better overview, we split the related work into two groups: \textbf{perturbations} and \textbf{concept validity}.

The impact of text manipulation on the resulting embeddings has been a subject of research for some time.
One of the most relevant to this paper is the one by Wang et al.~\cite{wangSNCSEContrastiveLearning2022}.
They make a distinction between different types of automated negations of text to train an embedding.
As such, it is closely related to both the Mission Impossible Languages by Kallini et al.~\cite {kalliniMissionImpossibleLanguage2024} method and GPT understanding by Liu et al.~\cite{liuGPTUnderstandsToo2024}.
All these methods alter input text to train LLMs; the latter two (by Kallini and Liu) focus on GPT, the first focuses on embedding.
Our method also has a focus on embeddings, yet we do not train on the generated data.
Furthermore, we also insert terms to create noise to compare with the negations.


Concept validation of embeddings in NLP has been studied in various contexts, including semantic similarity, interpretability, and alignment with human judgments.
One of the studies in this area was conducted by Fang et al.~\cite{fangEvaluatingConstructValidity2022}.
In this study, they describe how they alter and measure the ESS questionnaire texts.
They use a manual approach to generating texts, which ought to be more or less similar, describing the same properties.
The difference between similar and dissimilar is then used as the basis for a boxplot.
In our research, we do not alter any text manually, nor do we annotate the expected outcome.
We base ourselves purely on grammar, thus reducing bias in the generated output.
This is not to say that our method has no bias, but its impact is expected to be reduced~\cite{drorCognitiveHumanFactors2020}.

\section{Methods}
\begin{figure}
    \centering
    \includegraphics[scale=0.7]{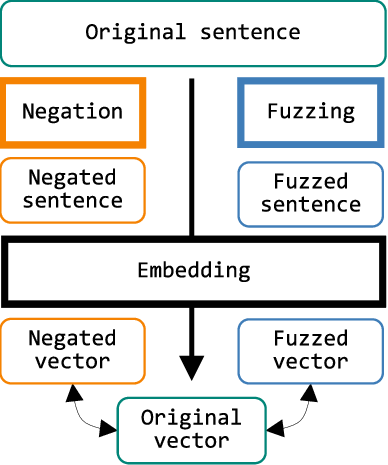}
    \caption{Approach setup, square components are algorithmic processes. This setup summarises the pipeline: from text alteration to embedding and similarity computation.}
    \label{fig:architecture}
    \vspace{-10pt}
\end{figure}

To measure whether a language model's capability of discerning between concepts, we introduce CSCs.
CSCs quantify how sensitively an embedding model reacts to semantic variations (meaning changes) compared to non-semantic variations (surface-level perturbations). This allows us to assess whether embeddings preserve conceptual meaning when sentence form changes.
The curves visualise the difference between the two types of alteration.
Specifically, each sentence is modified through two parallel processes: Fuzzing, which introduces non-semantic variations, and Negation, which introduces semantic variations. 
In this research, we use the terms Fuzzing and Negation as this makes the distinction between the different alterations easier to tell apart.
Our framework is not limited to strict Fuzzing and Negating, nor is the Negation a full opposite of the original sentence.
The embeddings of the altered and original sentences are then compared, and when applied across an entire corpus, the resulting distribution patterns form the CSCs.
The overview of the process is shown in Figure~\ref{fig:architecture}.

\subsection{Data}
The data used in this paper is intended to show the broad applicability of our method.
Furthermore, we wish to evaluate whether our method indeed works across domains and languages.
To evaluate our method, we aim to reduce the impact of other variables.
The two variables we wish to isolate in terms of effect are language and sentence length.
The amount of data available in a language has been shown to impact the quality of embedding algorithms~\cite{kotoZeroshotSentimentAnalysis2024}.
Hence, we use sources from two languages with differences in the amount of available content: Dutch and English.
For estimating the differences in available content, we used the number of available Wikipedia articles as a heuristic~\cite{wikisizes}.
This source was selected as it is commonly used for training many of the state-of-the-art LLM encoders.
In terms of difference, at the time of writing, the number of English pages is $62.3*10^6$ with a total of $48.0*10^8$ words~\cite{wikiEngStat}. 
This contrasts with the Dutch language, totalling $4.68*10^6$ pages and $5.30*10^8$ words~\cite{wikiDutchStat}. 
Such a large difference between the languages might be reflected in the performance of the algorithms.
The sentence length might also significantly impact the results.
As previously mentioned, the research by Liu et al.~\cite{liuGPTUnderstandsToo2024} shows the impact of inserting a singular term.
In our method, we also describe the process of adding a word to a sentence.
The impact in terms of volume strongly depends on the number of words present.
Altering a singular word in a five-word sentence can have a larger impact than on a twenty-word sentence.
We selected the corpora based on finding and detecting any of these issues.

The corpora we selected are CompetentNL (CNL), ESS Questionnaire (ESS), and Paracrawl (PC)~\cite{banonParaCrawlWebScaleAcquisition2020}.
CNL is a short sentence Dutch corpus describing skills.
For this research, we define a short sentence corpus as one where the majority of the sentences contain fewer than 10 tokens.
The ESS is the same as used by Fang et al.~\cite{fangEvaluatingConstructValidity2022}, it comprises a set of English questions of varying length.
PC is a corpus of both Dutch and English texts from web pages.
These crawled texts have a large variety of sentence lengths.
To give a better overview of the statistics per source, an overview is given in  Table~\ref{tbl:data_input}.
The main difference between PC and the others is that it is not as strongly bound to a singular domain.
CNL limits itself to short skill descriptions, and ESS specifically to questions from a singular questionnaire.
Hence, both can be labelled as narrow datasets.
The PC, however, is selected not to test on the domain, but to serve as a test of both language and sentence length.
The relationship between the sources regarding the main identified variables is depicted in Table~\ref{tbl:data_source}.
In this Table, we also introduce the PC\_filtered set.
This is a subset of the PC dataset, which is reduced to sentences with the same length as the CNL dataset.
This combination of sources is expected to give a good impression of the described variables.

\begin{table}
\centering
\begin{tabular}{c|cc}
                                   & \multicolumn{2}{c}{Sentence length} \\
                                   & Short              & Long           \\ \hline 
  Dutch   & CNL, PC\_filtered  & PC             \\
                           English & PC\_filtered       & ESS, PC
\end{tabular}
\caption{Overview of the different data sources used and key properties.}
\label{tbl:data_source}
\end{table}

\begin{table}
\resizebox{\columnwidth}{!}{%
\begin{tabular}{c||c|c|c|c}
\# Tokens & CNL     & ESS     & PC\_EN  & PC\_NL  \\ \hline \hline
0-10  & 99.26\% & 5.32\%  & 32.99\% & 33.41\% \\ \hline
10-20 & 0.72\%  & 30.85\% & 31.74\% & 31.21\% \\ \hline
20-30 & 0.02\%  & 26.60\% & 18.92\% & 18.71\% \\ \hline
30-40 & 0.00\%  & 26.60\% & 8.96\%  & 8.85\%  \\ \hline
40-50 & 0.00\%  & 10.64\% & 3.73\%  & 4.05\%  \\ \hline
$\ge$50   & 0.00\%  & 0.00\%  & 3.66\%  & 3.77\% \\ \hline\hline
Sentences & 4738    & 94    & 2560472 & 2560472
\end{tabular}

}
\caption{This table shows, per source, the percentage of sentences with a token count in a given range. Below the percentages, the total number of sentences from which the tokens were extracted is shown.
}
\label{tbl:data_input}
\vspace{-10pt}
\end{table}

\subsection{Fuzzing and Negation}\label{ssect:fuzzandneg}
Each of the sentences from the sources goes through two modification processes: Fuzzing and Negation, as shown in Figure~\ref{fig:architecture}.

\textbf{Fuzzing} serves as a control condition that introduces minimal, non-semantic textual perturbations, allowing us to test how stable an embedding remains when surface form changes but meaning is preserved. We define Fuzzing as the alteration of a sentence without altering its concept.

\textbf{Negation}, in contrast, represents a targeted semantic perturbation: it minimally changes the surface form while significantly altering the sentence’s conceptual meaning. This allows us to examine whether embeddings are sensitive to genuine semantic shifts rather than superficial textual ones.

Furthermore, we limit the Negation to change an equal number of tokens as the Fuzzing. This restriction on the Negation is to ensure that any measured effect can only be explained by the contents of the change, not by differences in the number of inserted tokens.

In this research, we implement both Fuzzing and Negation through token addition.
For Fuzzing, we insert articles in the respective languages, "de" or "het" in Dutch, and "a" or "the" in English. 
This simple insertion strategy aligns with many embedding methods that operate at the token level ~\cite{attention_is_all_you_need, devlin2019bert}.
Accordingly, we use a single-token insertion to introduce controlled surface-level variation.
For Negation, we insert a single negative particle: "niet" in Dutch and "not" in English. 
This addition minimally changes the sentence’s structure while reversing its propositional meaning.

\begin{figure}[h]
    \centering
    \includegraphics[width=0.5\linewidth]{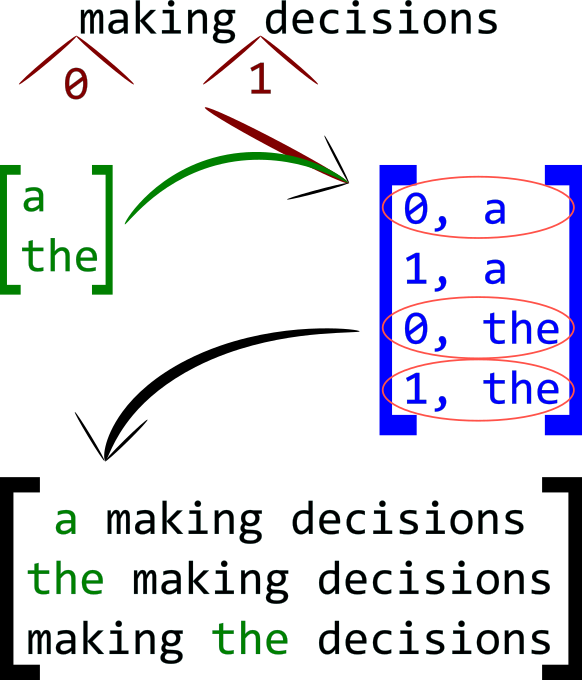}
    \caption{Depiction of the sentence generation process for the Fuzzing. Parts not visualised are the random shuffling. This sentence is translated from "beslissingen maken" from the CompetentNL source.}
    \label{fig:sentence_generation}
    \vspace{-10pt}
\end{figure}

The insertion procedure is visualised in Figure~\ref{fig:sentence_generation}.
The full procedure works as follows:
First, all viable locations for insertion are detected (displayed in red under the text).
Viable locations are defined as in front of any word in a sentence.
Second, given the available insertion terms (in green), create a list of all possible combinations of terms and locations (the list in blue).
This combination is randomly shuffled.
Finally, select and perform up to X options (red circles surrounding items in the blue list).
Each option results in a new sentence (as depicted at the bottom of the figure).
This procedure was chosen to be able to limit the data generation.
Short sentences do not support the same number of possible perturbations.
As such, longer sentences could be overrepresented in the curves if the X value is set too high.
There is also the increase in computations for higher X values.
As the impact of X is described by: $2X*n$ with $n$ being the size of the corpus.
Combining these factors, for our research we chose an X value of 3.
This data generation procedure works for both Fuzzing and Negation.

Although this algorithm is identical for Negating and Fuzzing, the number of sentences it returns for both is not guaranteed to be the same.
As depicted in Figure~\ref{fig:sentence_generation}, the insertion locations are base on the sentence and as such do not differ between Negation and Fuzzing.
The insertable items (shown in green in the figure) could differ between the operations.
Thus, the possible number of generated sentences (based on the length of the list in blue) could differ as well, given that each pair corresponds to an output sentence.
This difference is handled by normalisation in the comparison step (explained in subsection~\ref{ssect:Comparison}). 
For now, it is important to note that the chosen Fuzzing and Negation, although identical in their algorithm, are not identical in the expected output volume.

\subsection{Models}
In this research, we focus on understanding embedding methods.
Any model fitting the capabilities of turning a text into a vector (as depicted in Figures~\ref{fig:architecture} and \ref{fig:summary}) can be used.
As such, a non-Dutch model could be applied to the Dutch language.
However, such a mismatch ought to result in worse outcomes. 
To test this, we apply every model to every source, even if there is a mismatch between the model's supported language and the source language.

Firstly, for a baseline, we use \textbf{Term Frequency Inverse Document Frequency (TFIDF)}~\cite{scikit-learn}.
This is a simple yet effective method which does not consider word order and is not trained on a large corpus.
It is common to use a stopword removal step before applying this algorithm.
Yet, for our method to work, it is critical not to remove stopwords which might be added in the Fuzzing or Negation steps.
Therefore, instead of curating a stopword list, we decided to omit the stopword removal altogether.

The second approach we use is \textbf{Fasttext}~\cite{joulinBagTricksEfficient2016}.
This model is available in both Dutch and English in a pre-trained format.
Although it is an older model, it is a method which can embed sentences, making it a suitable baseline for embedding-based methods.

Finally, we use multiple \textbf{state-of-the-art sentence embeddings}; GroNLP~\cite{devries2019bertje}, MPNET~\cite{songMPNetMaskedPermuted2020}, RobBERTa~\cite{delobelleRobBERTDutchRoBERTabased2020} and LaBSE~\cite{fengLanguageagnosticBERTSentence2022}.
These models have been pre-trained on different sources.
GroNLP is specifically trained on Dutch corpora, although it might contain some English terms.
The MPNET and RobBERTa models have been trained on English corpora, while LaBSE is designed for cross-lingual applications using a variety of languages.
All of these models are based on the original BERT~\cite{devlinBERTPretrainingDeep2019}, yet each is configured or trained differently to suit specific needs.

\subsection{Concept Separation Curves}\label{ssect:Comparison}
The goal of Concept Separation Curves is to illustrate the understanding of a text embedding model without annotations.
This method does so by generating at least three vectors for one text: the original text vector, the fuzzed vector, and the negated text vector.
The hypothesis for our method centres around two different observables:
\begin{itemize}
    \item The Fuzzed Vectors should stay close to the original embedding. The meaning is the same, but a change was made regardless. 
    \item The Negated Vectors should show a difference with the original, which is larger than the Fuzzed Vector differences. The meaning differs from the original; as such, the impact on the vector ought to be greater.
\end{itemize}

To perform this comparison, we propose \textbf{Concept Separation Curves} (CSCs).
CSCs are a visualisation of how a text embedding technique responds to concept alteration compared to textual alteration.
The response to concept alteration and textual alteration impact curves are both plotted in the same graph.
These curves are made by comparing the vectors of the original sentence with both the negated and fuzzed vectors.
This results in two curves, one for the negations and one for the fuzzing.
These curves are intended to show the sensitivity to Fuzzing and how different the Negation impacts the resulting vectors.
To improve the readability of the curves and focus on the underlying distribution, we perform a Gaussian kernel density estimation function~\cite{2020SciPy-NMeth} on the raw data.
The resulting curves can differ wildly as the Negation and Fuzzing process do not guarantee an equal output volume (as explained in subsection~\ref{ssect:fuzzandneg}).
To this end, we perform the surface normalisation as defined in equation~\ref{eq:norm}.

\begin{equation}
    norm(d, i) = \frac{d_i}{\sum_{j=-1}^{1} d_j}
    \label{eq:norm}
\end{equation}

In this normalisation, let $d$ represent the density and $i$ the inspected value within the range [-1,1], and $d_i$ the density at $i$.
Just like $i$, $j$ also iterates over the values in $d$, meaning that for both there is an overarching resolution.
The meaning of this resolution is the number of steps to inspect in this [-1, 1] range.
Intuitively, this normalisation ensures that the total area under each density curve equals 1, allowing a fair comparison between the Negation and Fuzzed distributions regardless of their differing volumes.
The resulting values are plotted in a line plot for both the negated and fuzzed densities.

One aspect which can be difficult to spot in the plot is the shared surface between the curves.
Due to possible small differences across all similarities, these differences can add up without being easy to spot.
To circumvent this, we also compute the shared surface using the equation shown in equation~\ref{eq:results}.

\begin{equation}
    \sum_{i=-1}^1{min(norm(fuz, i), norm(neg, i))}
    \label{eq:results}
\end{equation}

In this equation, $fuz$ stands for the fuzzed similarity density distribution and $neg$ for the negated.
The function describes how we normalise these density distributions of $f$ and $n$, resulting in each adding up to 1.
The normalisation is required to compensate for the potential difference in volume.
This difference in volume is expected due to potential differences in negated terms and fuzzed terms (as explained in subsection~\ref{ssect:fuzzandneg}).
Then we are interested in the overlap between these values, and go through values $i$ from -1 to 1, taking the minimum value between $fuz$ and $neg$.
This results in an overlap score ranging from 0 to 1, where 0 indicates no overlap and 1 is a perfect overlap.

Finally, there is a chance that some alterations in the vector are due to the text being out of distribution.
Because our generated sentences are not expected to always have correct grammar, it might be that embedding models like BERT generate wildly different vectors.
To check for this, we use a method based on Wang et al.~\cite {wangSNCSEContrastiveLearning2022} to generate negations with more grammatically correct negations.
In their original method, they add a negation if none is present and add one if not present.
We adjust this by only allowing the addition.
Furthermore, this method is specific to the English language. 
Thus, we only apply it to the English corpora.

\section{Results}
In this section, we present the main patterns observed in the CSCs, illustrating both desirable and failure-case behaviours, followed by a quantitative analysis of their overlap.
First, we start with an expected and desired curve, followed by different types of negative results.

\begin{figure}[h]
    \centering
    \includegraphics[width=0.9\columnwidth]{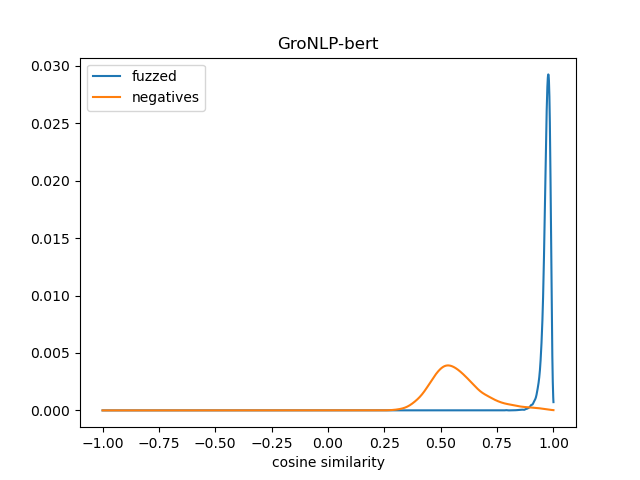}
    \caption{Concept Separation Curves using Gaussian kernel density estimation on the CNL data and the GroNLP embedding model. This graph shows an overlap of 0.0221.}
    \label{fig:cnl_gronlp}
\end{figure}

An example of good concept separation is shown in Figure~\ref{fig:cnl_gronlp}. 
In this figure, CNL data is used in combination with the GroNLP model.
This plot shows a nice separation between concepts.
The fuzzed curve is close to 1, thus showing a high similarity to the original.
The negated sentence curve is to the left of the fuzzed curve, showing an increased dissimilarity to the original sentence.
As such, it displays that the encoded vector governs a different concept.
The overlap is indicative of two groups: fuzzed sentences where the Fuzzing had a large impact, and Negations with a small impact.
As this overlap is small and the Negations are more dissimilar to the original compared to the fuzzed sentences, this displays a good concept separation.

\begin{figure}[h]
    \centering
    \includegraphics[width=0.9\columnwidth]{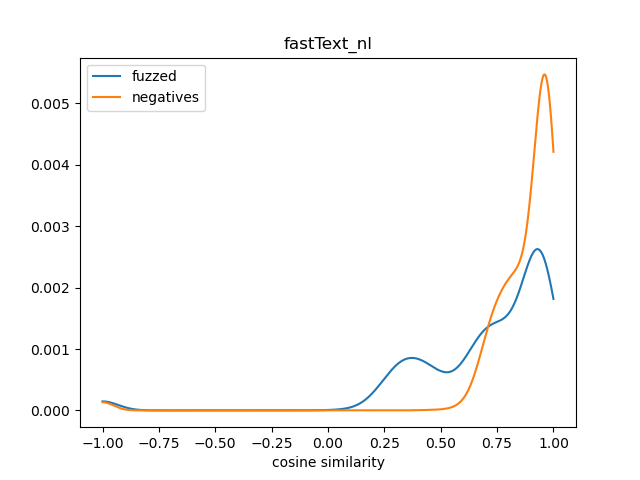}
    \caption{CNL data with FastText embedding. The overlap is 0.6652}
    \label{fig:cnl_ft}
    \vspace{-10pt} 
\end{figure}

\begin{figure}[h]
    \centering
    \includegraphics[width=0.9\columnwidth]{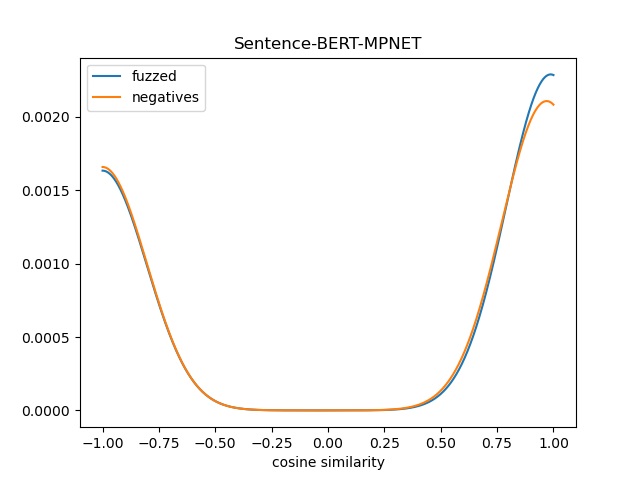}
    \caption{ESS data with sBERT MPNET embedding. The overlap is 0.9810}
    \label{fig:ess_mpnet}
    \vspace{-10pt} 
\end{figure}

Not all embeddings perform as nicely as the already shown GroNLP.
For instance, the FastText embedding, as shown in Figure~\ref{fig:cnl_ft}.
Here, the Negations are more similar to the original sentence than the fuzzed sentences.
As such, this algorithm cannot be stated to have encoded the concept of a sentence.
Another example of a less-than-desirable result is the one shown by sBERT MPNET in Figure~\ref{fig:ess_mpnet}.
Although the fuzzed sentences are slightly more similar to the original, there is a second peak near -1.
This can be seen across datasets and languages.
Given that the only constant in our alteration is the addition of a token, we have to conclude that this algorithm reacts heavily to the change of token position.
As such, it behaves more akin to a hashing algorithm rather than a concept embedding.
Both these patterns show that the used embedding has difficulty discerning concepts. 

\begin{figure}
    \centering
    \subfigure[Unfiltered, 0.5168]{\includegraphics[width = 0.465\columnwidth]{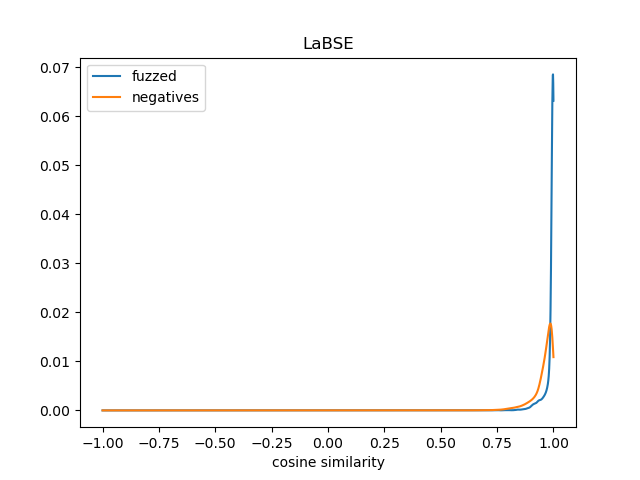}} 
    \subfigure[Filtered, 0.4632]{\includegraphics[width = 0.465\columnwidth]{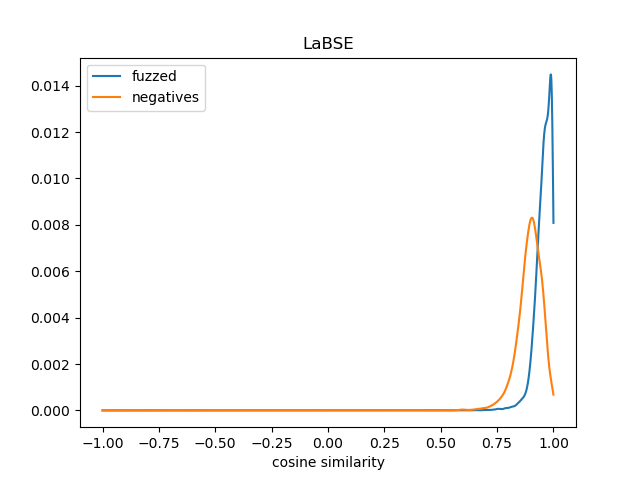}}
    \caption{LaBSE algorithm on PC NL data, in both its raw form and filtered to the short sentence lengths present in the CNL dataset.}
    \label{fig:pc_labse_impact}
    \vspace{-10pt} 
\end{figure}

\begin{table*}[htp]
\begin{tabularx}{\textwidth}{l|XXXXXX}
            & CNL            & PC NL Filt.    & PC EN Filt.    & PC NL          & PC EN             & ESS  \\ \hline
TFIDF       & 0.2993         & 0.7684         & 0.6015         & 0.9449         & 0.8614            & 0.4516  \\
GroNLP      & \textbf{0.0221}& \textbf{0.0861}& 0.8056         & \textbf{0.2925}& 0.9449            & 0.8767  \\
LaBSE       & 0.1984         & 0.4632         & \textbf{0.3588}& 0.5168         & \textbf{0.5635}   & \textbf{0.3511}  \\
Fasttext NL & 0.6652         & 0.6825         & 0.8809         & 0.7402         & 0.8692            & 0.8072  \\
Fasttext EN & 0.7764         & 0.7977         & 0.7652         & 0.8804         & 0.9405            & 0.9549  \\
MPNET       & 0.9756         & 0.9813         & 0.9496         & 0.9874         & 0.9523            & 0.9810  \\
RobBERTa    & 0.9605         & 0.9726         & 0.9488         & 0.9905         & 0.9410            & 0.9773   \\

\end{tabularx}
\caption{Normalised area overlap of Fuzzed and Negated curves.}
\label{tbl:RESULTS_B}
\end{table*}

The final pattern of note we discovered in our results is the sentence length effect.
This effect is visible in Figure~\ref{fig:pc_labse_impact}. 
The unfiltered and filtered data are from the same domain and language.
The only difference is the length of the sentences.
The curves of the unfiltered, therefore longer texts, are less pronounced and thus more difficult to perceive. 
With even longer sentences, this effect is expected to worsen.
As such, we see that with increased sentence length, our method decreases in its detection capabilities.

In previously shown results, the overlap is included in each plot.
The complete overview is given in Table~\ref{tbl:RESULTS_B}.
In this Table, there are stark differences between models.
The Dutch GroNLP appears to be the best at each Dutch dataset, and the same is the case for LaBSE in English.
Furthermore, some models appear to have a near-perfect overlap as the values are close to 1.
This holds for longer sentences as well, even though optical inspection of the curves for this data is more difficult.

\section{Discussion}
In this paper, we demonstrated a method for measuring the concept certainty of embedding algorithms.
Unlike other approaches, our approach does not rely on human annotations, nor does it utilise a classifier.
This makes our method easy to apply in other languages and even makes it accessible per domain.

\paragraph{Effects of Sentence Length and Token Position.}
When inspecting the results, the length of a sentence appears to be the strongest limiting factor.
This makes sense as we alter a smaller percentage compared to the size of the sentence.
Altering in a significant manner for such text would either involve larger insertions or shortening the sentences first.
The problem with the larger insertions is that with each insertion, some meaning could be altered.
For instance, in short sentences, there is already some impact with the Fuzzing, as there might be a difference between "a bike" and "the bike".
This effect is cumulative; thus, when applied in larger volumes, the concept is nearly bound to change.
As such, we expect that it might be beneficial for our method to reduce sentences to a shorter format.

Next to the sentence length, some algorithms struggled with the position of words.
Especially in short sentences, the MPNET and RobBERTa algorithms changed the vectors drastically for both Negation and Fuzzing.
Whilst the impact of Fuzzing and Negation ought to be different, they were nearly negligible.
We believe this is due to the position of the tokens in the sentence.
These results are more akin to a positional hashing algorithm, rather than a content embedding.
Given that both insertion algorithms do not append at the end of a sentence, this might result in the findings for both algorithms.

\paragraph{Implications for Embedding Evaluation.}
Finally, the impact of language on the task appeared to be present, yet minor.
When inspecting Table~\ref{tbl:RESULTS_B} and the Figures, a pattern emerges that certain models perform best for a specific language.
Other than the best, there is a difference between the languages in terms of the spread of overlap.
When looking at the statistics between the English data and Dutch data, we see that Dutch has an average of 0,6668 ($\sigma$ 0,2658) and English 0,7992 ($\sigma$ 0.1622).
This difference is not completely fair, as there are more English models than Dutch models in our test.
However, it is a rather striking difference, which could be explained by differences between languages.

\section{Conclusion}
In this research, we looked into a method for finding the concept validity of embedding models.
Our method encompassed a difference analysis of Fuzzing and Negation to accomplish this.
We found that there are differences between models in terms of validity. 
The proposed Concept Separation Curves visualise and quantify how much a sentence embedding cares about meaning. 
Furthermore, some models were unable to distinguish between the same and different concepts, whereas others were not.
Additionally, we found that some algorithms reacted heavily to the position of tokens. 
Given the range of applied datasets, we expect this effect is not tied to the used data in terms of length and language.
As such, it appears these models do not capture the underlying concepts.

\section{Limitations}
A potential limitation of the present study is its reliance on the incorporation of terms into a sentence.
However, it should be noted that this method may not be universally applicable, as in some languages, addition cannot be used to obtain a negative value.
Consequently, a potential avenue for future research could involve exploring alterations as a substitute for additions

An additional constraint is that the available resources for the languages do not encompass antonyms and synonyms.
Instead of inserting terms for both Fuzzing and Negation, it would be possible to use synonyms and antonyms for these operations.
An effort was made to use publicly available antonyms and synonyms. 
However, analysing the number of synonyms and antonyms in our dataset showed that, on average, each sentence had fewer than 1 word from either, with a negligible number of sentences containing both a synonym and an antonym.
As such, it was deemed not viable, for it would not alter enough sentences.
If a more complete set of synonyms and antonyms were available, this would be of interest to compare against the obtained results from this research.

An important note to the results presented in this research is that our measure is not directly related to its capabilities in other tasks.
The simplest example would be a list comprising all search terms in a domain.
When looking for an item, it can be retrieved quickly and precisely, even while the user is still typing the full query. 
Such a system would not have high concept validity.
As the Fuzzing might completely offset its results, a precise position might be relevant.
Thus, our research does not indicate the performance of other tasks 
It can be used, however, in domains where certainty plays a key role.
The implications of our research for the results in other fields, however, are a topic which might warrant further research.

Another point of interest is the extension of the proposed method.
In the results, we highlighted different stereotypical curves. 
These effects are not guaranteed to be visible from the total overlap.
An example would be an embedding where the Fuzzing has a peak near -1 and a Negation near 1.
Such a curve would display that the method can make a distinction between concepts and correctly have an overlap value of 0.
However, the problem would be that there might be some underlying effect causing such an abnormality.
Possibly due to token order sensitivity and not filtering the negated words.
We did not encounter this effect, but the method could be extended to include measurements to check for such an effect.

The measurement employed, the cosine similarity, is another potential avenue for enhancement.
The usage of the cosine similarity reduces the comparison to a single number, whilst the true change of the perturbation might be too local for a significant change.
As such, a model with a higher dimensionality (for instance, 700+), reacting to our method on just 1 dimension, would perform worse compared to a lower-dimensional model.
To correct for this, a more thorough comparison of the vectors could be devised.

Finally, it is possible to extend our method to different types of input alteration.
These alterations could involve altering a noun or verb, compared to changing a domain-specific irrelevant term.
An example would be: "Applicant should be able to drive a car", Fuzzing it into "He should be able to drive a car" and the alteration being "Applicant should be able to drive a boat".
The problem we found with such alterations is that they are more domain-bound and require more complex algorithms to automatically construct (if at all possible).
The alterations themselves could be done in a more fine-grained approach tailored to a specific domain.

\bibliographystyle{lrec2026-natbib}
\bibliography{other_work.bib}

\section*{Appendix}\label{sec:add}
In this appendix, we show the additional curves underlying our conclusion. 
The results displayed in this section are grouped by data source.
For each source, the figures are plotted in a reduced format to illustrate the overall curve.
The full-size figures are appended at the end of the paper.
Finally, the table with all computed overlap scores is shown.

\begin{figure*}
    \centering
    \subfigure[Fasttext]{\includegraphics[width = 0.32\linewidth]{images/output/UWV_fastText_nl.png}} 
    \subfigure[GroNLP]{\includegraphics[width = 0.32\linewidth]{images/output/UWV_GroNLP-bert.png}}
    \subfigure[LaBSE]{\includegraphics[width = 0.32\linewidth]{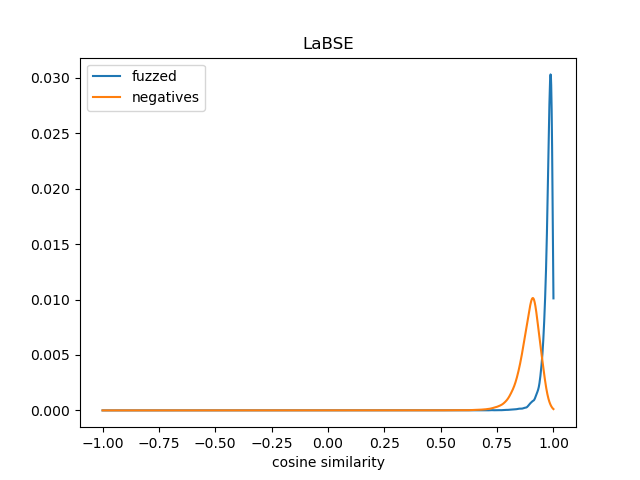}}
    
    \subfigure[sBERT MPNET]{\includegraphics[width = 0.32\linewidth]{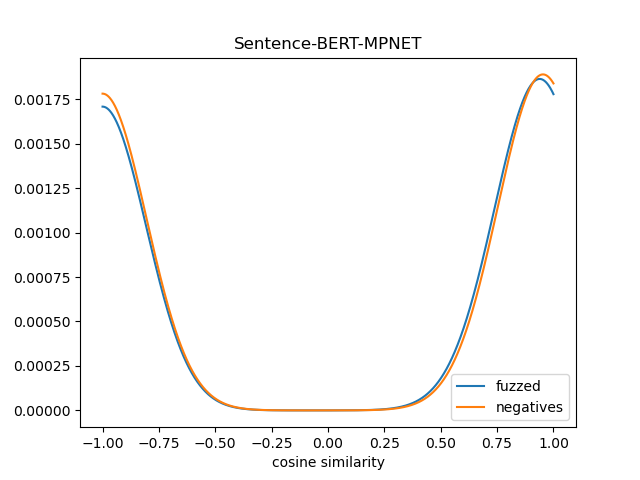}} 
    \subfigure[sBERT RobBERTa]{\includegraphics[width = 0.32\linewidth]{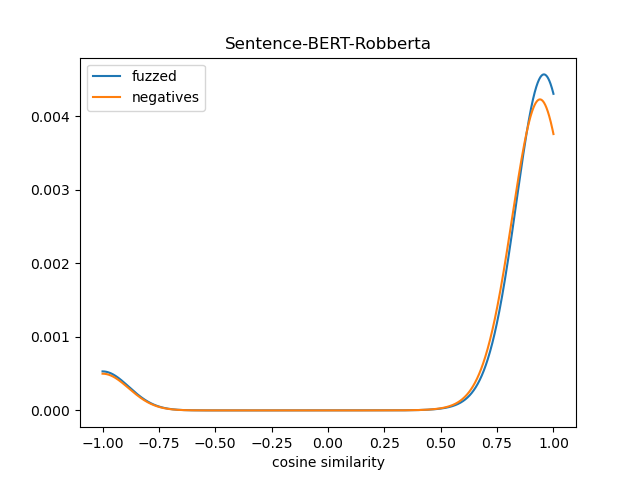}}
    \subfigure[TFIDF]{\includegraphics[width = 0.32\linewidth]{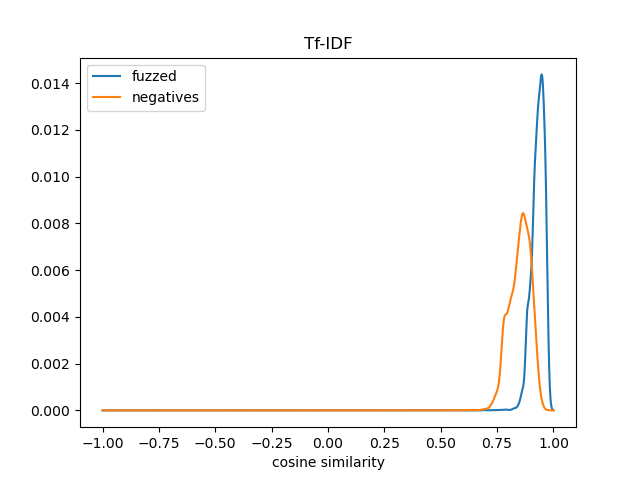}}
    
    \caption{CNL results}
    \label{fig:uwv_results}
\end{figure*}

First, we look into the CNL results.
When comparing the different embedding methods in Figure~\ref{fig:uwv_results}, it is apparent that specific models underperform significantly and TFIDF works better than expected.

\begin{figure*}
    \centering
    \subfigure[Fasttext]{\includegraphics[width = 0.32\linewidth]{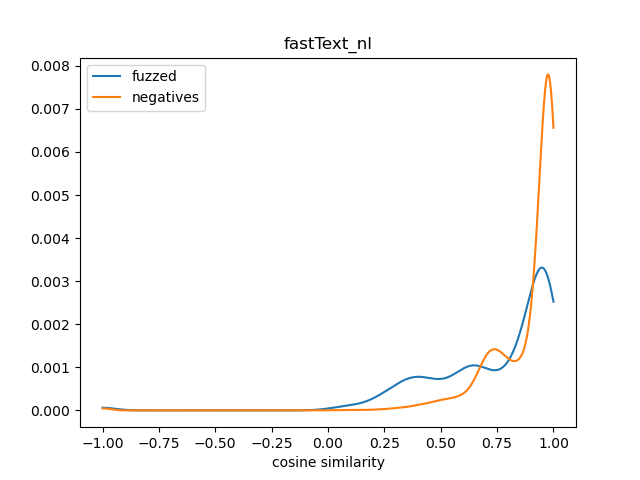}} 
    \subfigure[GroNLP]{\includegraphics[width = 0.32\linewidth]{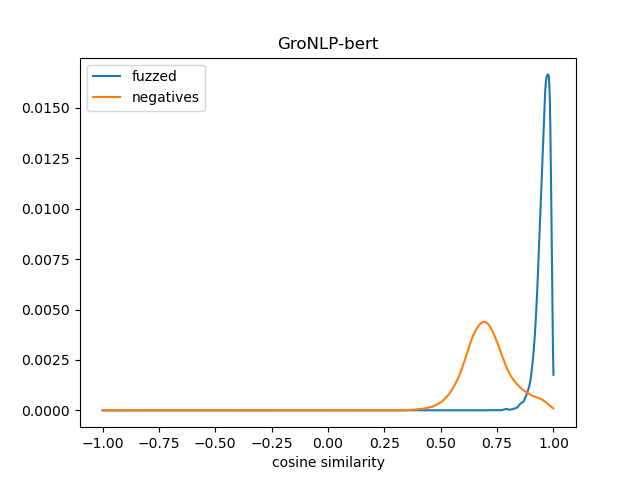}}
    \subfigure[LaBSE]{\includegraphics[width = 0.32\linewidth]{images/output/filtered/Paracrawl_nl_LaBSE.png}}
    
    \subfigure[sBERT MPNET]{\includegraphics[width = 0.32\linewidth]{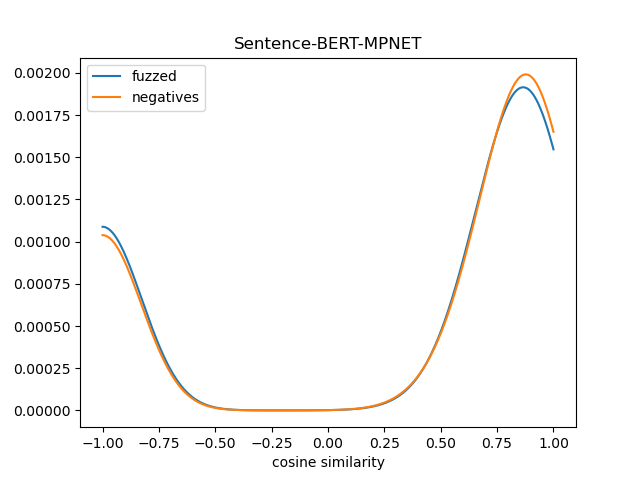}} 
    \subfigure[sBERT RobBERTa]{\includegraphics[width = 0.32\linewidth]{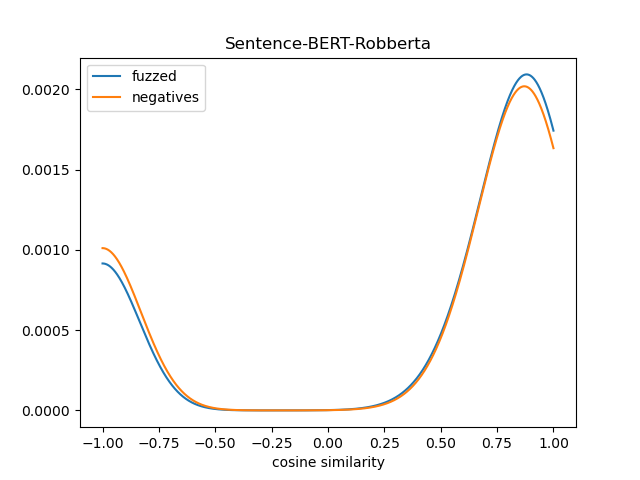}}
    \subfigure[TFIDF]{\includegraphics[width = 0.32\linewidth]{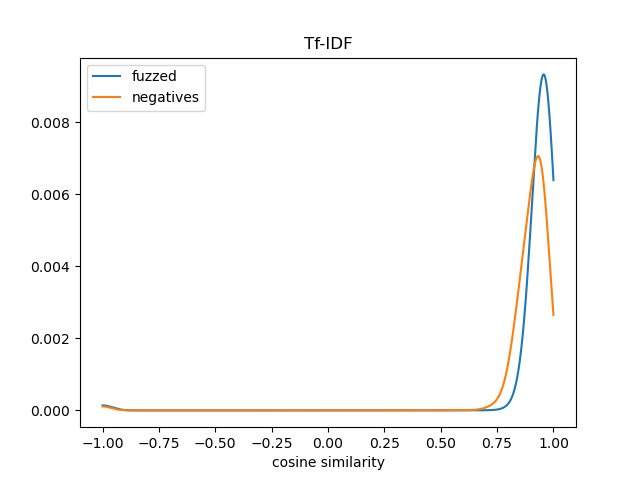}}
    
    \caption{PC NL filtered results}
    \label{fig:pc_filtered_results}
\end{figure*}

The shortened Paracrawl data is displayed in Figure~\ref{fig:pc_filtered_results}.
It strongly resembles the CNL results, yet the exact curves per algorithm are different.
Specifically, the MPNET, RobBERTa and TfIDF show diverging results.

\begin{figure*}
    \centering
    \subfigure[Fasttext]{\includegraphics[width = 0.32\linewidth]{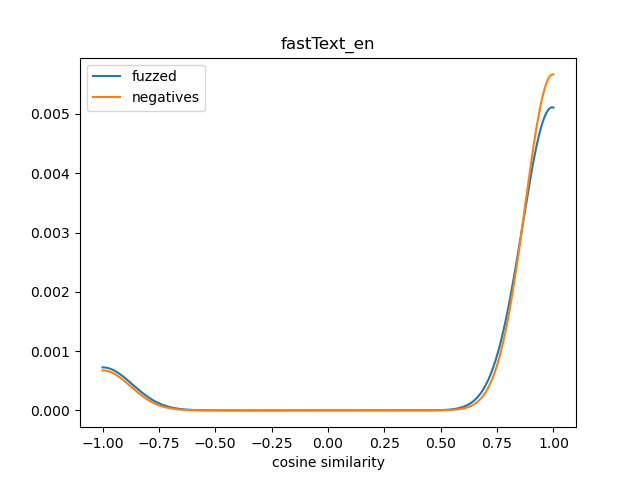}} 
    \subfigure[GroNLP]{\includegraphics[width = 0.32\linewidth]{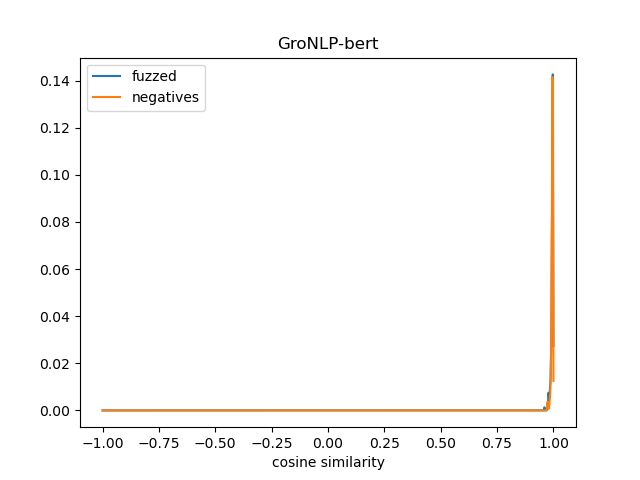}}
    \subfigure[LaBSE]{\includegraphics[width = 0.32\linewidth]{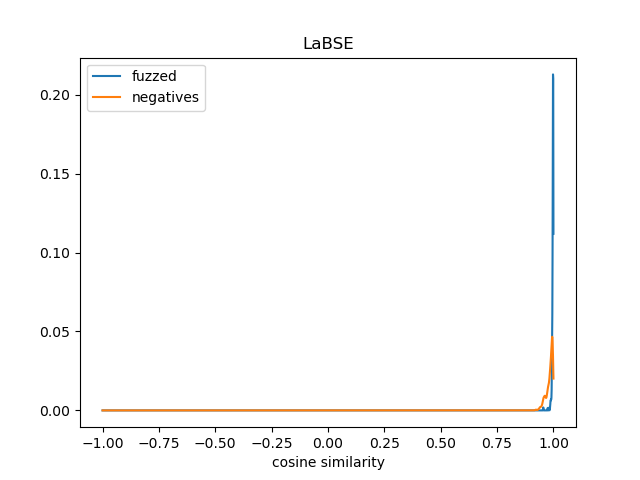}}
    
    \subfigure[sBERT MPNET]{\includegraphics[width = 0.32\linewidth]{images/output/Qixian_Sentence-BERT-MPNET.png}} 
    \subfigure[sBERT RobBERTa]{\includegraphics[width = 0.32\linewidth]{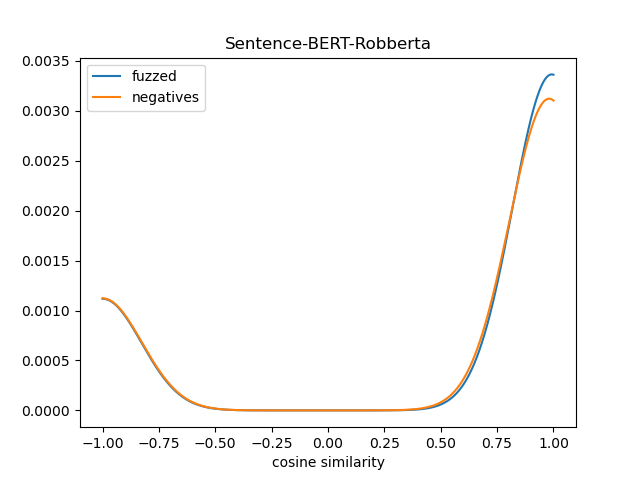}}
    \subfigure[TFIDF]{\includegraphics[width = 0.32\linewidth]{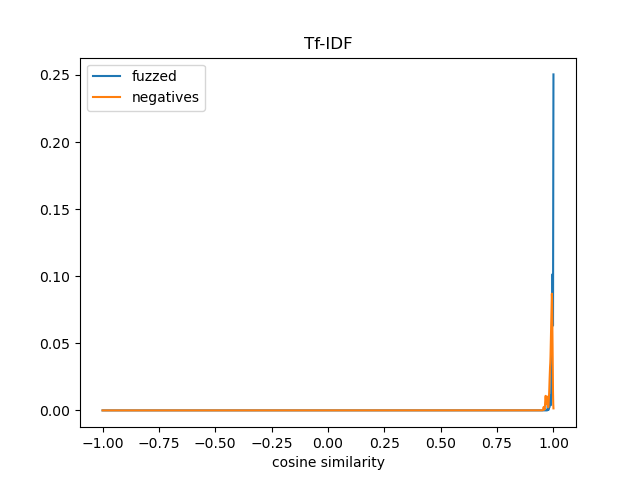}}
    
    \caption{ESS results}
    \label{fig:ess_results}
\end{figure*}

The ESS questionnaire results shown in Figure~\ref{fig:ess_results} show a strong reduction in the impact of negated and fuzzed.
There are differences in the fuzzed and negated curves, but they lack the pronounced differences shown in previous figures.

\begin{figure*}
    \centering
    \subfigure[Fasttext]{\includegraphics[width = 0.32\linewidth]{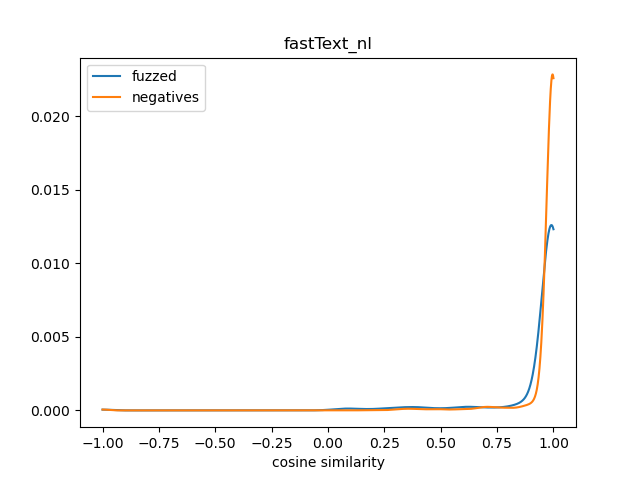}} 
    \subfigure[GroNLP]{\includegraphics[width = 0.32\linewidth]{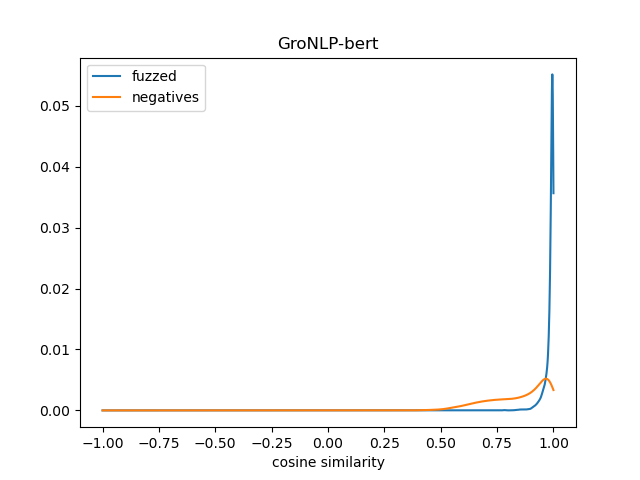}}
    \subfigure[LaBSE]{\includegraphics[width = 0.32\linewidth]{images/output/unfiltered/Paracrawl_nl_LaBSE.png}}
    
    \subfigure[sBERT MPNET]{\includegraphics[width = 0.32\linewidth]{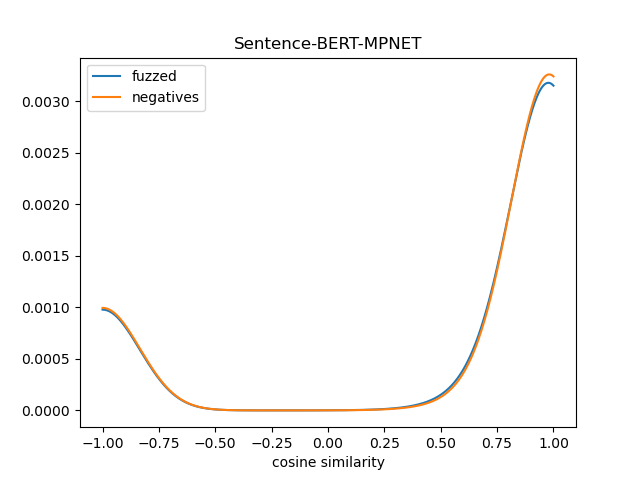}} 
    \subfigure[sBERT RobBERTa]{\includegraphics[width = 0.32\linewidth]{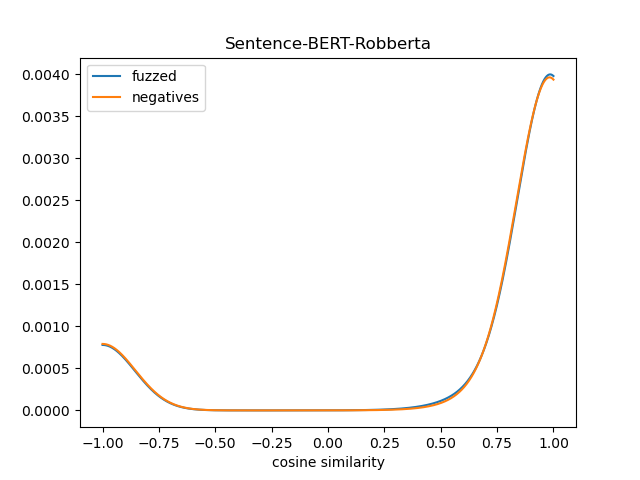}}
    \subfigure[TFIDF]{\includegraphics[width = 0.32\linewidth]{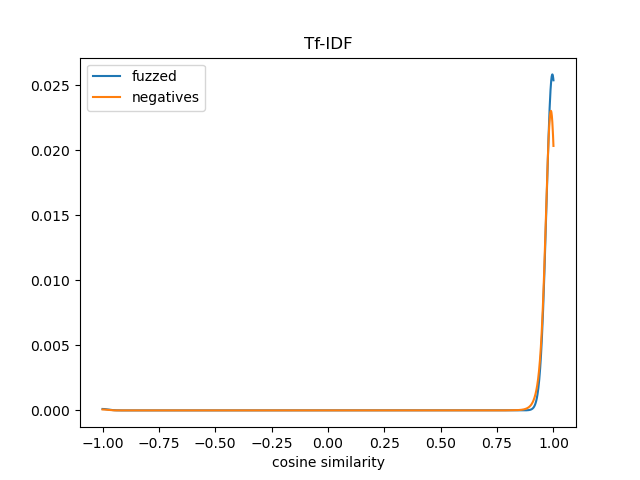}}
    
    \caption{PC NL unfiltered results}
    \label{fig:pc_unfiltered_results}
\end{figure*}

The full-length Paracrawl set is displayed in Figure~\ref{fig:pc_unfiltered_results}.
Here, the effects of the alteration are also relatively small compared to one of the short sentence formats.
As such, it follows the same pattern as the ESS questionnaire data.

\end{document}